\pdfoutput=1

\documentclass[11pt]{article}

\usepackage[]{acl}

\usepackage{times}
\usepackage{latexsym}
\usepackage{graphicx}
\usepackage{amsmath}
\usepackage{bbm}

\usepackage[T1]{fontenc}
\usepackage{url}
\usepackage{hyperref}
\usepackage{xurl}

\usepackage[utf8]{inputenc}

\usepackage{microtype}
\usepackage{pgfplots}
\usepackage{booktabs}
\usepackage{framed}
\usepackage{algorithmic}
\usepackage{multirow}
\usepackage{xspace}
\usepackage{fontawesome5}
\usepackage{xcolor}

\usepackage{caption}
\usepackage{subcaption}

\usepackage{inconsolata}
\usepackage{booktabs}
\usepackage{multirow}
\usepackage{soul}
\usepackage{enumitem}
\usepackage[colorinlistoftodos,prependcaption,textsize=tiny]{todonotes}

\newcommand{\dsname}{\textbf{\texttt{KIWI}}\xspace}

\definecolor{darkgreen}{RGB}{54,128,45}
\definecolor{forestgreen}{RGB}{152.9,255.0,77.5}
\definecolor{limegreen}{RGB}{85.4,140.3,130.5}

\definecolor{NavyBlue}{HTML}{1E88E5}
\definecolor{OrangeRed}{HTML}{D81B60}
\definecolor{Gold}{HTML}{dbcd09}

\newcommand{\turboShort}{\raisebox{-2pt}{\includegraphics[height=1em]{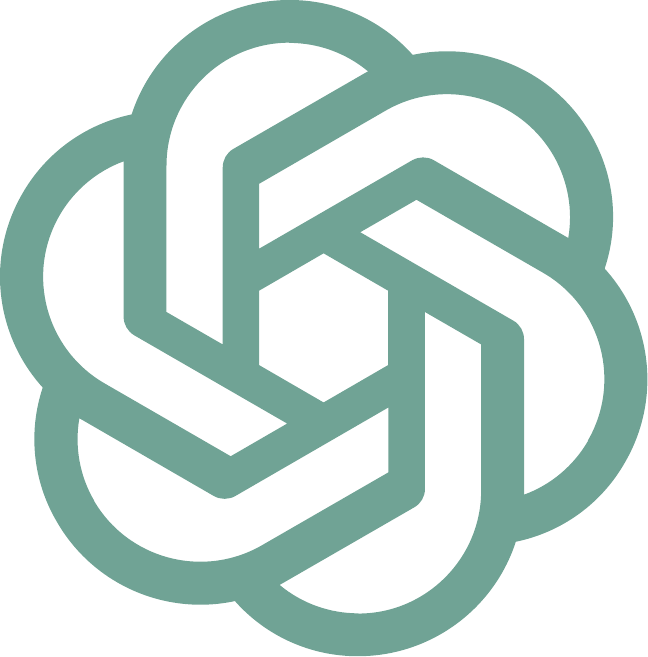}} \texttt{gpt-3.5}\xspace}
\newcommand{\turboIcon}{\raisebox{-2pt}{\includegraphics[height=1em]{icons/openai.pdf}}}
\newcommand{\turboName}{{\texttt{gpt-3.5} \xspace}}

\newcommand{\gptFourFull}{\raisebox{-2pt}{\includegraphics[height=1em]{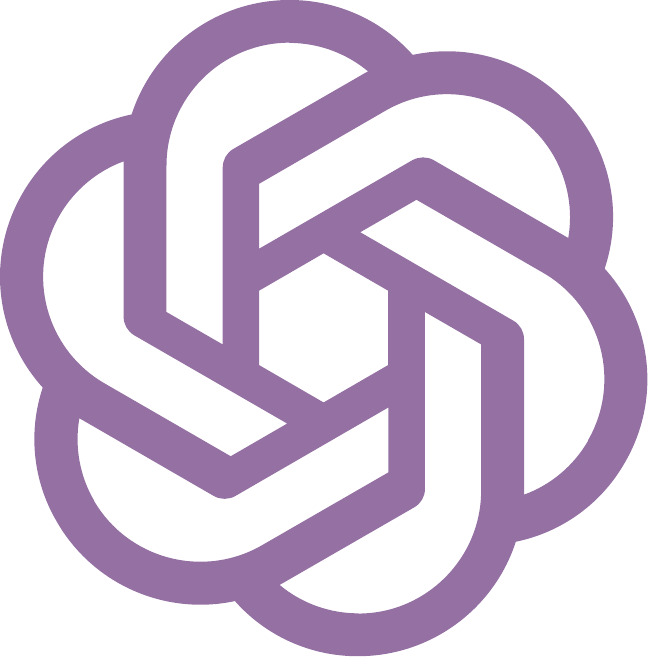}} \texttt{gpt-4}\xspace}
\newcommand{\gptFourShort}{\raisebox{-2pt}{\includegraphics[height=1em]{icons/gpt4.pdf}} \texttt{gpt-4}\xspace}
\newcommand{\gptFourIcon}{\raisebox{-2pt}{\includegraphics[height=1em]{icons/gpt4.pdf}}}
\newcommand{\gptFourName}{{\texttt{gpt-4} \xspace}}

\newcommand{\llamachatShort}{\raisebox{-2pt}{\includegraphics[height=1em]{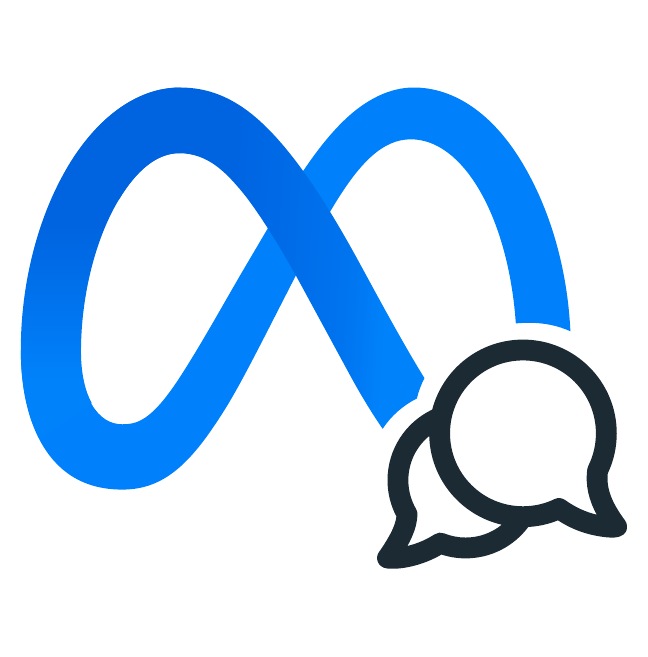}} \texttt{llama2}\xspace}
\newcommand{\llamachatIcon}{\raisebox{-2pt}{\includegraphics[height=1em]{icons/llama2chat.pdf}}}
\newcommand{\llamachatName}{{\texttt{llama2} \xspace}}

\newcommand{\kiwiIcon}{\raisebox{-2pt}{\includegraphics[height=1em]{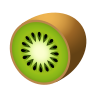}}}

\title{\kiwiIcon \dsname: A Dataset of Knowledge-Intensive Writing Instructions\\
for Answering Research Questions}

  \author{Fangyuan Xu$^{\diamondsuit}$\thanks{$^{*}$Work performed during an internship at AI2.}~~  Kyle Lo$^{\heartsuit}$ ~~ Luca Soldaini$^{\heartsuit}$ ~~ Bailey Kuehl$^{\heartsuit}$ ~~ Eunsol Choi$^{\diamondsuit}$~~ David Wadden$^{\heartsuit}$\\
$^\diamondsuit$The University of Texas at Austin, $^\heartsuit$Allen Institute for AI\\
 \hspace{0.5em} {\texttt{\{fangyuan, eunsol\}@utexas.edu}} \\
 {\texttt{\{kylel, lucas, baileyk, davidw\}@allenai.org}}
 }

\begin{document}
\maketitle
\begin{abstract}

Large language models (LLMs) adapted to follow user instructions are now widely deployed as conversational agents. In this work, we examine one increasingly common instruction-following task: providing writing assistance to compose a long-form answer. 
To evaluate the capabilities of current LLMs on this task, we construct \dsname, a dataset of knowledge-intensive writing instructions in the scientific domain. Given a research question, an initial model-generated answer and a set of relevant papers, an expert annotator iteratively issues instructions for the model to revise and improve its answer. We collect 1,260 interaction turns from 234 interaction sessions with three state-of-the-art LLMs. 
Each turn includes a user instruction, a model response, and a human evaluation of the model response.  
Through a detailed analysis of the collected responses, we find that all models struggle to incorporate new information into an existing answer, and to perform precise and unambiguous edits. Further, we find that models struggle to judge whether their outputs successfully followed user instructions, with accuracy at least 10 points short of human agreement. Our findings indicate that \dsname will be a valuable resource to measure progress and improve LLMs' instruction-following capabilities for knowledge intensive writing tasks.\footnote{Our dataset is released at \url{https://www.cs.utexas.edu/~fxu/kiwi/}.}

\end{abstract}

 \begin{figure*}[t!]
 \vspace{-2em}
    \centering
    \includegraphics[trim={0 0 0 0},clip,width=\textwidth]{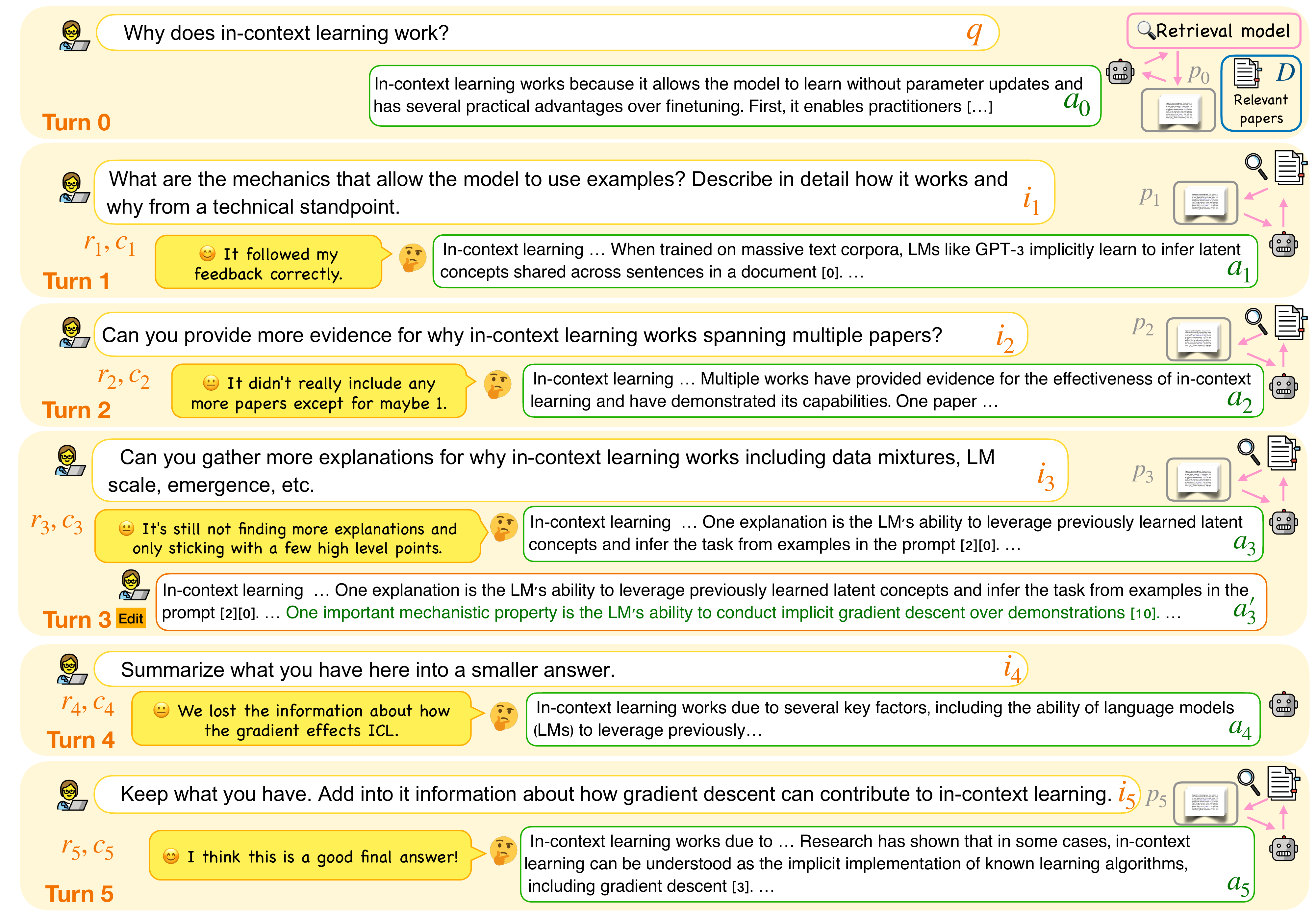}
    \caption{An example interaction session between a user and our system. Given a question $q$ and a set of documents $\mathbf{D}$, the model first generates an initial answer $a_0$. At each turn $t$, the user issues an instruction $i_{t}$, for which an updated answer $a_{t}$ is generated. The user provides a rating $r_{t}$ and a comment $c_{t}$ for the generated answer, and optionally edits the answer (such as $a_{3}'$ in turn 3) such that the edited answer completely follows the instruction.}
    \label{figs:intro}
\end{figure*}

\section{Introduction}

As LLM-powered conversational agents \cite{openai2023gpt, Touvron2023Llama2O, Anil2023GeminiAF} have gained widespread adoption, users have turned to these systems for assistance on a wide range of real-world tasks. In particular, recent works studying LLM interactions ``in the wild'' \cite{Ouyang2023TheSA, Zheng2023LMSYSChat1MAL} have discovered that 10-20\% of user queries contain requests for writing assistance, i.e. using an LLM to create, revise, or organize a piece of written text. However, we currently lack an understanding of LLMs' capabilities as writing assistants, {especially in knowledge-intensive settings}. What types of writing instructions do users issue to LLMs, and how well can LLMs follow different types of instructions? 

We present \dsname, a dataset of expert-written \textbf{K}nowledge-\textbf{I}ntensive \textbf{W}riting \textbf{I}nstructions to better understand LLMs' instruction-following abilities as writing assistants. To collect \dsname, we set up an interactive interface between a researcher and an LLM (Figure \ref{figs:intro}). We first prompt an LLM to generate a long-form answer \cite{fan-etal-2019-eli5} to an NLP research question, based on a set of passages from relevant research papers. A researcher then iteratively issues instructions for the model to revise the answer and evaluates the model-generated revision. The interaction session continues until the user is satisfied with the final answer or a maximum number of turns is reached. \dsname contains instructions collected from 234 interaction sessions with three state-of-the-art language models (GPT-4,GPT-3.5-turbo and LLaMA-2-70b-chat), providing a total of 1,260 unique instruction instances. Each instance consists of (1) a user instruction, (2) the model's previous and (3) revised answers, and (4) a human annotator's judgment (both categorical rating and free-form explanation) on whether the revised answer followed the instruction.

Using \dsname, we conduct an in-depth analysis to characterize the types of instructions issued by researchers, and to measure how well models can follow different types of instructions. We find that LLMs \textit{do not} excel at this task yet, with the best model (GPT-4) achieving success for only 59\% of the instructions. Specifically, LLMs fail to precisely follow user's instructions (such as satisfying location and length constraints), often cause answer quality to degrade when integrating new information into a previous answer, and struggle to avoid making changes to answers that are not requested.

Finally, we examine the ability of the strongest LLM (GPT-4) to evaluate whether a response follows an instruction, comparing its judgement to the user ratings in \dsname. Unlike prior works which found that LLMs can judge outputs as reliably as humans for open-ended instruction following \cite{Zheng2023JudgingLW, alpaca_eval}, our experiments show that GPT-4 cannot reliably evaluate responses for instructions in \dsname, which are often specific and precise. While the best performing model which retrieves and prepends 10-shot examples improves upon zero-shot performance by 5\% accuracy, it still lags behind human agreement by 12\% accuracy. Given these findings, we believe that \dsname will be a valuable resource for future research in instruction following and evaluation. 

\section{Interaction Design}\label{sec:task_setup}

\subsection{Design goals}
Unlike prior works \cite{Kopf2023OpenAssistantC,zhao2023wildchat, Zheng2023LMSYSChat1MAL} which deploy an LLM in the wild to collect diverse types of user interaction data, we focus on the use case of seeking writing assistance. We design a knowledge-intensive interaction setting with two key considerations. First, the information in the answer should be grounded to a set of relevant documents. This allows us to examine model's ability to leverage information from \textit{multiple} documents, motivated by real-world writing tasks such as literature review \cite{Shen2023BeyondSD}. Second, the user interacts with the model for \textit{multiple} turns to iteratively revise the answer. Instead of collecting a single editing instruction for (question, initial answer) pairs, this setting not only adheres to the iterative nature of the text editing process \cite{Collins1980AFF,Du2022UnderstandingIR}, but also allows us to collect diverse and fine-grained instructions across different editing stages as we will see in \S \ref{subsec:overview}.

\subsection{Interaction process} \label{subsec:annot_process}
We provide an overview of the interaction process (Figure \ref{figs:intro}) and describe the data collection procedure in \S \ref{sec:data_collection}. The system inputs are an NLP research question $q$ and a collection $\mathbf{D}$ of research papers which, collectively, contain information sufficient to write a high-quality, paragraph-length answer to $q$. To begin the interaction process, an LLM generates a ``first draft'' answer $a_0$ to $q$ based on $\mathbf{D}$. Each interaction turn $t$ involves four steps with actions taken by the user (\faUser) or the model (\faLaptop): (1) The user reads the LLM-generated answer and issues an instruction $i_t$ to the model to revise the answer. (2) The model then generates an updated answer $a_{t}$ given $q$, $a_{t-1}$ and $i_{t}$. (3) The user provides a rating $r_{t}$ and a comment $c_{t}$ evaluating how well $a_{t}$ follows $i_{t}$. (4) The user can choose to edit the answer $a_{t}$ if it did not follow their instruction, producing $a_{t}'$. Afterwards, the user can start a new turn by issuing another instruction for $a_{t}$ (or $a_{t}'$). The interaction continues until either the user is satisfied with the answer, or a total of 10 turns is reached. 
We describe each step below with additional details (prompts, hyperparameters) in Appendix \ref{sec:system_details}.

\paragraph{\faLaptop\xspace Generating the initial answer} Given the question $q$, we first retrieve a set of five passages from $\mathbf{D}$ with a retrieval  model $R$, denoted as $R(q, \mathbf{D})$. We prepend the retrieved passages to the question $q$ to generate the initial answer $a_{0}$ with a language model $M$, that is, $a_{0}$ = $M(q, R(q, \mathbf{D}))$.\footnote{We perform retrieval because prepending the full text of multiple papers (each with around 6,000 tokens) exceeds the context window of the LLMs (with at most 8,000 tokens).}

\paragraph{\faUser\xspace Issuing instructions} To start a turn $t$, the user issues an instruction $i_{t}$, specifying how they would like the LLM to revise its answer in the previous turn $a_{t-1}$, and categorizes its type $I_{t}$. We identify two types of instructions. \textbf{Information-seeking} instructions require the model to incorporate new information into the answer; e.g. Turn 3 of Figure \ref{figs:intro} asks the model to acquire additional information from the source documents $\mathbf{D}$. \textbf{Stylistic} instructions ask the model to transform content that is already present; e.g. Turn 4 in Figure \ref{figs:intro} asks the model to compress its answer. For each interaction turn, the user indicates the type of the instruction $I_{t}$ by selecting a checkbox in the annotation interface.

\paragraph{\faLaptop\xspace Answer revisions} Next, the model $M$ is prompted to revise its previous answer $a_{t-1}$ to satisfy user instruction $i_t$. We use one prompt for information-seeking instructions (which require passage retrieval from $\textbf{D}$), and another for stylistic questions (which do not).

For information-seeking instructions, we first leverage a retrieval model $R$ to retrieves five passages $p_{t}$ from $\mathbf{D}$, using the instruction $i_t$ as the query. Next, we prompt $M$ to summarize the information in $p_t$ that is likely to be relevant given ($a_{t-1}$, $i_t$); we denote this summary as $s_t$. Finally, we generate an updated answer by prompting $M$ with ($q$, $a_{t-1}$, $i_{t}$, $s_{t}$).\footnote{In preliminary experiments, we found that this ``summarize-then-answer'' workflow yielded better results than directly prompting $M$ with the retrieved passages $p_{t}$.} For stylistic instructions, we generate the updated answer directly by providing ($q$, $a_{t-1}$, $i_{t}$) as input. We only include a single previous answer in the prompt to generate the updated answer; in the pilot study, we found that this approach yielded comparable performance to including the entire interaction history.

\paragraph{\faUser\xspace Measuring answer quality}

After a revised answer $a_{t}$ is generated, the user provides a rating indicating whether $M$ successfully followed their instruction: $r_{t} = \mathbf{r}(i_{t}, q, a_{t-1}, a_{t})$. We instruct the annotators to rate $a_{t}$ as one of the following:

\begin{itemize}[leftmargin=*,itemsep=0pt,topsep=3pt,partopsep=0pt,parsep=0pt]
    \item \texttt{good}: Successfully followed the instruction. 
    \item \texttt{neutral}: Partially followed the instruction. 
    \item \texttt{bad}: Didn't follow the instruction at all, or modified the answer to an undesirable state. 
\end{itemize}

Since our goal is to measure the helpfulness of the model response as determined by the annotator, we ask annotators to use their own judgement in selecting a rating rather than providing a detailed guideline.\footnote{To alleviate annotator burden, annotators are not required, but welcomed, to check the factual correctness of model-generated answers. Assessing faithfulness of LLM generations represents an important but orthogonal research direction.} Nonetheless, we find in \S \ref{sec:exps} that annotators have high agreement on this categorization. We also collect a mandatory free-form comment $c_{t}$ from the annotator explaining their rating $r_t$. Example turns are in Table \ref{tab:rating_examples} in the Appendix \ref{subsec:data_examples}.

\paragraph{\faUser\xspace Handling incorrect instruction-following} When $r_{t} \neq $ \texttt{good}, answer $a_{t}$ does not fully follow instruction $i_t$ and needs to be corrected. While the user might issue instructions in the subsequent turns to correct what has gone wrong with $a_{t}$, 
 
the success of the user's correction attempt depends heavily on the model's instruction following ability. 
The user might fall into a loop of continuing to prompt a model which is not able to perform the action requested in the instruction. To avoid such scenarios, we allow, but do not require, the user to edit $a_{t}$ when $r_{t} \neq $ \texttt{good}, which induces $a_{t}'$, such that $\mathbf{r}(i_{t}, q, a_{t-1}, a_{t}') = $ \texttt{good}. If the user chooses to not edit the answer and $r_{t} =$ \texttt{bad}, we remove the answer from the answer history when generating answers in the subsequent turns.

\section{Dataset Collection}\label{sec:data_collection} 

We first describe our approach to curate high-quality $(q, \mathbf{D})$ pairs (\S\ref{subsec:question_annotation}), followed by our annotation protocol for interaction collection (\S\ref{subsec:annot_details}).

\subsection{(Question, Document set) creation}\label{subsec:question_annotation}

The process proposed in \S \ref{subsec:annot_process} requires high-quality $(q, \textbf{D})$ pairs to serve as a starting point for interaction collection. We aim to create $(q, \textbf{D})$ pairs which satisfy the following desiderata: (1) the questions should be realistic and likely to be asked by a researcher, (2) the questions should be challenging, requiring information from multiple documents and (3) each document set $\textbf{D}$ should contain sufficient information to answer its corresponding question. We observe that the related work section of a research paper often answers a set of implicit research questions, and supplies the documents answering each question as citations. Thus, we collect high-quality $(q, \textbf{D})$ pairs by writing questions implied by related work sections, and pairing each question with its cited papers.  

\paragraph{Source article selection} We select a set of roughly 100 papers published in ACL 2023 across 11 different tracks listed in the ACL handbook.\footnote{We choose to focus on NLP research questions, since the authors of this paper have expertise in this domain.} This set of papers covers a wide range of NLP research topics and was published after the training data cutoff time of the language models used for data collection in this work. The model thus has to answer the question based on the documents, instead of memorizing from its training data.

\paragraph{Question creation and filtering} Three of the authors with prior NLP research experience manually annotate 88 questions from 75 papers. Given a source paper, the annotator first reads through its related work section and decides if a good question can be derived. A good question should fulfill the following criteria: (1) it should ``stand alone'', i.e. it is not anchored in the source paper; and (2) there should be at least four articles cited in the related work which 
contain relevant information for answering the question.
If a good question can be derived, the annotator writes a question $q$, together with the list of evidence papers $\mathbf{D}$ and the section of the related work $r$, from which the question is derived. We query the S2ORC \cite{lo-etal-2020-s2orc} corpus to retrieve the full text of each paper and extract paragraphs. To ensure sufficient coverage, we filter out questions with either (1) more than one evidence paper missing in S2ORC or (2) fewer than 4 evidence papers in total. We obtain a total of 78 questions with an average of 6 cited papers per question.

\paragraph{Analysis of collected questions} Our annotated questions consist of 12 words on average, with the majority of questions starting with "How" (47\%) and "What" (46\%). Upon manual examination, all models are able to generate a reasonable initial answer given the question and retrieved passages. This shows that our pipeline creates high quality data: the document set indeed contains sufficient information to answer the question.\footnote{In the initial phase of the project, we explored using GPT-4 to generate questions given related work paragraphs, but found that the generated questions were often not answerable by the cited papers. We include analysis on question generation in the Appendix \ref{sec:question_generation}.} See example questions in Table \ref{tab:example_questions} in the Appendix \ref{subsec:data_examples}.

\begin{table}
\footnotesize
\begin{center}
\begin{tabular}{p{7.5cm}}
\toprule

\textbf{Source Paper}: Small pre-trained language models can be fine-tuned as large models via over-parameterization \cite{Gao2023SmallPL}\\ 
\midrule
\textbf{Related work paragraph}: Over-parameterization in Neural Network. Over-parameterization has shown the superiority on providing better model initialization (Arpit and Bengio, 2019), improving model convergence (Du et al., 2019; Allen-Zhu et al., 2019b; Gao et al., 2022a) and generalization (Allen-Zhu et al., 2019a). [...] \\
\textbf{Question}: Why does over-parameterization lead to improved initialization and convergence of deep neural networks? \\ 
\textbf{Evidence papers}: \citet{Gao2021AGC, AllenZhu2018ACT, Li2016TernaryWN, Du2018GradientDP} \\

\bottomrule
\end{tabular}
\caption{An example ($q$, $\mathbf{D}$) annotation (\S \ref{subsec:question_annotation}). Given a source paper, the annotator first selects the related work paragraph(s) that imply the question, then writes the question and extracts evidence papers cited.}
\end{center}
\end{table}

\subsection{Interaction data collection} \label{subsec:annot_details}
At the start of an interaction session, the annotator is shown the question, initial answer, and the titles of relevant papers for the question. During the interaction, the annotator is shown the full interaction history, and the retrieved passages from the last info-seeking instruction. We release annotation instruction\footnote{\url{https://docs.google.com/document/u/1/d/e/2PACX-1vTrsM5r-p5kjy6Ue7AkbLqwOmJ-GoK9kwToBEnW-Nwk00yd4tKSFkWu9p63j7rH-PvNwwNXLzTyRWhi/pub}} and provide annotation interface screenshot as Figure \ref{fig:annotation_interface} in Appendix \ref{sec:system_details}.

\paragraph{Language models} We conduct annotations with three models $M$: GPT-3.5-turbo, GPT-4 and LLaMA-2-chat(70B) \cite{Touvron2023Llama2O}, competitive commercial and public LLMs available at the time of data collection, respectively. For all models, we decode with temperature 0.7. During data collection, the identity of model $M$ is not revealed to the annotator. For each question, we collect one interaction session with each model. We ensure that annotators do not answer the same question more than once. 

\paragraph{Retrieval models} We use the state-of-the-art dense retriever Contriever \citep{izacard2021contriever} finetuned on MSMARCO ~\citep{Campos2016MSMA} as our retriever $R$. \footnote{In preliminary experiments, we found that Contriever outperformed sparse methods like BM25 \cite{Robertson2009ThePR}, which struggle at handling synonyms.}

\paragraph{Annotators} To ensure our annotators have expertise in answering research questions, we recruit NLP researchers through Upwork\footnote{\url{https://www.upwork.com/}} and professional networks. We pay the annotators USD \$25 per hour on average. The annotators first participate in a paid pilot study, which involves reading the instructions and completing an annotation example. A total of 22 annotators participated in our pilot annotations and 15 of them performed final annotations for \dsname. The final group of annotators consists of four people with a Ph.D. degree, seven NLP Ph.D. students,  one person with a Master's degree and three undergraduate students. All annotators have at least one year of NLP research experience. On average, each interaction session required 15 minutes to complete.
 To ensure that the annotators are familiar with the subject matter of the question, we first collect their preferences on a set of topics derived from the track information of each question's source paper. We assign questions to annotators based on their preferences. The annotations were collected from July to December 2023. In total, our annotations cost \$2,240 USD.

\section{Analysis of \dsname} \label{sec:analysis}
We first present results on overall model performance (\S \ref{subsec:overview}). Then we conduct a fine-grained analysis to understand the different types of instructions (\S \ref{sec:instruction_analysis}) and errors made by LLMs (\S \ref{sec:error_analysis}).

\begin{table}
\begin{center}
\small
\begin{tabular}{@{}lccc@{}}
\toprule
\textbf{Model} & \textbf{\# turn (edited)} & \textbf{avg. \# t/s} & \textbf{\% info/style}\\\midrule

\gptFourName &  370 (55) & 4.7 & 59/41 \\ 
\turboName & 402 (111)  & 5.2 & 61/39 \\ 
\llamachatName & 488 (149) & 6.3 & 44/56 \\
\midrule
\textbf{Total} & 1,260 (315) & 5.4 & 54/46 \\ 
\bottomrule
\end{tabular} 
\caption{Data statistics of the \dsname. We collect a total of 234 sessions (78 sessions per model). We report the total number of turns, number of turns with edited answers, number of turns per session, and the distribution of instruction types for each model.}\label{tab:data_stats}
\end{center}
\end{table}

 \begin{figure}[t!]
    \centering
    \includegraphics[width=0.48\textwidth]{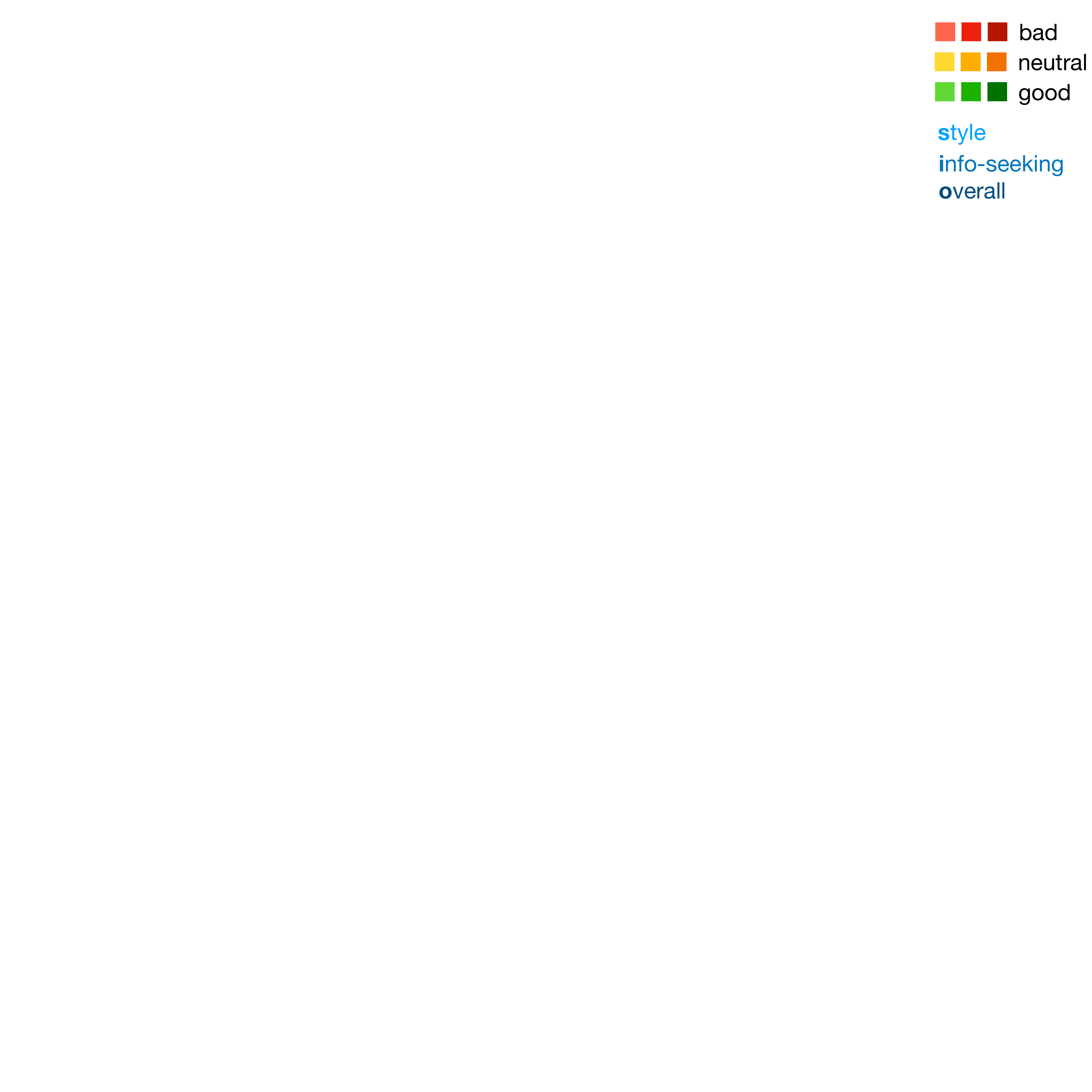}
    \caption{Top: distribution of annotator ratings. The left columns represent ratings for stylistic instructions (s), the middle columns for info-seeking instructions (i) and the right columns for overall (o) ratings. Bottom: distribution of user-issued instructions across the session.}
    \label{figs:feedback_type_across_session}
\end{figure}

\begin{table*}[]
\resizebox{0.99\textwidth}{!}{%
\begin{tabular}{@{}p{2cm}clp{6cm}@{\hspace{2ex}}p{7.5cm}@{\hspace{3ex}}@{}}
\toprule
\textbf{Type} & \textbf{\% info/style} & \textbf{\% good} & \textbf{Definition} & \textbf{Example Instruction} \\ 
\midrule
\multirow{3}{=}{More information (43\%) } & \multirow{3}{*}{\textbf{88}/12} & \gptFourIcon: \textcolor{NavyBlue}{\textbf{57}} & \multirow{3}{=}{Asking for auxiliary information related to the question or the answer (e.g. example, results, limitation)} & \multirow{3}{=}{How well do state-of-the-art models perform on these datasets?} \\
& & \turboIcon: \textcolor{NavyBlue}{45}&  &  \\
& & \llamachatIcon: \textcolor{orange}{34} &  &  \\
\midrule


\multirow{3}{=}{Expand coverage (9\%)} & \multirow{3}{*}{\textbf{86}/14} & \gptFourIcon: \textcolor{NavyBlue}{\textbf{40}} & \multirow{3}{=}{Asking for more information directly answering the question, usually from multiple papers.} & \multirow{3}{=}{Keep the text content as it is, but extract examples of applications of contrastive explanations to NLP from papers 0,1, and 2.} \\ 
& & \turboIcon: \textcolor{orange}{31} &  &  \\
& & \llamachatIcon: \textcolor{red}{12} &  &  \\
\midrule 

\multirow{3}{*}{Clarification (5\%) } & \multirow{3}{*}{\textbf{69}/31} & \gptFourIcon: \textcolor{NavyBlue}{\textbf{52}} & \multirow{3}{=}{Asking for clarification of a concept already in the answer} & \multirow{3}{=}{Do all of the quantization methods you mention require additional training?} \\ 
& & \turboIcon: \textcolor{NavyBlue}{50} &  &  \\
& & \llamachatIcon: \textcolor{NavyBlue}{50} &  &  \\

\midrule 

\multirow{3}{*}{Remove (13\%)}  & \multirow{3}{*}{4/\textbf{96}} & \gptFourIcon: \textcolor{darkgreen}{\textbf{77}} & \multirow{3}{=}{Requests to remove certain part of the answer} & \multirow{3}{=}{Remove the repeated techniques listed in the last paragraph.}  \\ 
& & \turboIcon: \textcolor{orange}{28} &  &  \\
& & \llamachatIcon: \textcolor{red}{16} &  &  \\
\midrule


\multirow{3}{*}{Specific edit (6\%)}   & \multirow{3}{*}{23/\textbf{77}} & \gptFourIcon: \textcolor{NavyBlue}{\textbf{59}} & \multirow{3}{=}{Requests to directly add information that are provided in the instruction verbatim to the answer} & \multirow{3}{=}{At the end of the final paragraph, add the sentence "Evaluation of factuality for natural text remain active research lines."} \\
& & \turboIcon: \textcolor{orange}{35} &  &  \\
& & \llamachatIcon: \textcolor{red}{19} &  &  \\
\midrule


\multirow{3}{*}{Condense (10\%)} & \multirow{3}{*}{4/\textbf{96}}  & \gptFourIcon: \textcolor{darkgreen}{\textbf{63}} & \multirow{3}{=}{Requests to shorten or summarize the answer} & \multirow{3}{=}{Please shorten the answer to a maximum of 6 paragraphs.}  \\ 
& & \turboIcon: \textcolor{orange}{27} &  &  \\
& & \llamachatIcon: \textcolor{orange}{23} &  &  \\
\midrule


\multirow{3}{*}{Reorganize (11\%)} & \multirow{3}{*}{13/\textbf{87}} & \gptFourIcon: \textcolor{darkgreen}{\textbf{74}} & \multirow{3}{=}{Requests to reorganize the answer (e.g. move things around)} & \multirow{3}{=}{Reorganize the answer so that positive results are described first, followed by challenges and open problems}  \\ 
& & \turboIcon: \textcolor{orange}{27} &  &  \\
& & \llamachatIcon: \textcolor{orange}{34} &  &  \\

\midrule

\multirow{3}{*}{Others (3\%)} & \multirow{3}{*}{\textbf{53}/47} & \gptFourIcon: \textcolor{darkgreen}{\textbf{62}} & \multirow{3}{=}{Edits that don't belong to any of the category above}  & \multirow{3}{=}{Don't change anything further about this answer but copy it as-is.} \\ 
& & \turboIcon:  \textcolor{orange}{23} &  &  \\
& & \llamachatIcon: \textcolor{orange}{29} &  &  \\
\bottomrule
\end{tabular}
}
\caption{Definitions and example instructions for the fine-grained instruction types (\S \ref{sec:instruction_analysis}). We report the distribution classified by GPT-4 and \% good rating (color coded) across the three models (\gptFourShort, \turboShort, \llamachatShort).} 
\label{tab:example_instructions}
\end{table*}

\subsection{Dataset overview}  \label{subsec:overview}
We present statistics of the interaction sessions in Table \ref{tab:data_stats}. We observe that the number of turns required to reach a satisfactory answer negatively correlates with model capability, with \gptFourName requiring the fewest turns and \llamachatName requiring the most. Comparing instruction types, around 60\% of instructions are information-seeking for \gptFourName, and 45\% for \llamachatName; this suggests that the \gptFourName is able to more quickly generate stylistically acceptable text, allowing users to focus on information content. We also observe that users tend to issue information-seeking instructions early in the interaction session followed by stylistic instructions (Figure \ref{figs:feedback_type_across_session}). This confirms that collecting instructions over multiple turns allows us to cover diverse instructions occurring in different writing stages.

\paragraph{\dsname presents a challenge for existing LLMs:} The distribution over user ratings of model responses is shown in Figure \ref{figs:feedback_type_across_session}. \gptFourName exhibits the strongest instruction-following performance (59\% good responses), while \llamachatName exhibits the weakest (27\% good responses). Interestingly, trends for the two types of instruction differ---while \gptFourName provides good responses to stylistic instructions around 70\% of the time, it struggles at following information seeking instructions (i.e. adding new information to an answer). In contrast, \llamachatName performs extremely poorly at following stylistic feedback, with only 15\% of responses rated as good. All models fail to follow user instructions in more than 40\% of responses, leaving significant headroom for LLMs' instruction-following ability.

\subsection{Fine-grained instruction analysis}\label{sec:instruction_analysis} \label{subsec:finegrained_instructions}

Based on our findings that LLMs struggle at following the instructions in \dsname, we conduct a fine-grained analysis to categorize the different types of instructions found in the dataset, and measure model performance on each. We manually label a sampled set of instructions, finding eight categories, and then scale up using GPT-4. We construct a few-shot prompt containing [(original question, instruction) $\rightarrow$ fine-grained instruction type] demonstrations for each instruction type, and prompt GPT-4 to label the entire dataset. Details about the prompt and human analysis can be found in the Appendix \ref{sec:instruction_analysis_auto}. We validate our approach by confirming that GPT-4 achieves 92.5\% agreement with humans on a held-out set of 40 instructions.

\paragraph{\dsname contains diverse instruction types:} The results of our analysis are shown in Table \ref{tab:example_instructions}. Although we did not reference coarse-grained category labels when defining the fine-grained categories, we observe that our resulting fine-grained categories can be divided based on our distinction between information-seeking and stylistic instructions. Information-seeking instructions include requests to provide {auxiliary} information, expand answer coverage, or provide clarification. Stylistic instructions range from lower-level requests such as performing a specific edit, to more complicated operations such as condensing and reorganizing the answer. We report additional statistics for each category in Table \ref{tab:edit_stats} in the Appendix \ref{sec:instruction_analysis_auto}.

\paragraph{Models struggle at integrating new information and precise editing:} Among the information seeking instructions, {``expand coverage''} is the most difficult, as this type of instruction requires both successful retrieval of relevant passages from \textit{multiple} papers and coherent integration into the previous answer. For stylistic instructions, \gptFourName performs well at removal and reorganization, achieving more than 70\% good ratings, while \turboName and \llamachatName perform poorly. Interestingly, all three models, including \gptFourName,  struggle to make specific edits (e.g. add a requested sentence verbatim), which are usually trivial for humans. While prior work has found that LLMs excel at single document text summarization \citep{goyalzeroshotnews2022}, we observe poor performance for the {``condense''} instructions (which request to shorten previous answer), with \gptFourName's response rated as good only 63\% of the time. This is perhaps because answers in our task involve information from \textit{multiple} papers, and multi-document summarization remains a challenge for current LLMs \cite{shaib-etal-2023-summarizing}.

\subsection{Error analysis}\label{sec:error_analysis}
To better understand \textit{how} models fail to follow instructions, we analyze the free-form comments $c_t$ written by the annotators. As in \S \ref{sec:instruction_analysis}, we manually label a small set of comments and scale up with GPT-4 to all 760 turns which received a \texttt{neutral} or \texttt{bad} rating. We construct a prompt containing [(instruction, comment $\rightarrow$ target category)] demonstrations for each fine-grained error category; see Appendix \ref{sec:comment_analysis_auto} for details of human analysis and the exact prompt. On a validation set with 22 examples, GPT-4 achieves 90\% agreement with humans.

\begin{table}
\begin{center}
\small
\resizebox{0.48\textwidth}{!}{
\begin{tabular}{@{}p{3cm}p{5.5cm}@{}}
\toprule
{\textbf{Error type}} & \textbf{Example} \\
\midrule
\textbf{Unrequested change} (31\%) \newline The update introduces changes not requested. & \textbf{Instruction}: Limit the number of sentences describing advantages and disadvantages for each method to 3. \newline \textbf{Comment}: The model shortened the desired paragraphs but added another and removed information from the previous answer. \\ 
\midrule
\textbf{Ignored} (21\%) \newline The requested change was not made. &  \textbf{Instruction}: Reduce the discussion about task-specific pretraining objectives \newline \textbf{Comment}: Model does not seem to have reduced any information at all. \\ 
\midrule 
\textbf{Constraint failure} (11\%) \newline The update in the answer does not follow some constraint(s). & \textbf{Instruction}: As first sentence of the text provide a definition of "clarification question".  \newline \textbf{Comment}: It gave the definition but not as first sentence in the text.	\\
\midrule
\textbf{Poor integration} (9\%) \newline The answer is less coherent after the update. & \textbf{Instruction}:  Please start the answer by describing what "leveraging future utterance for dialogue generation" is. \newline \textbf{Comment}: The requested info was added. However, the quality of the answer degraded somewhat and now has a odd ordering. \\  
\midrule
\textbf{Others} (28\%) \newline  The update is not satisfactory for other reason. & \textbf{Instruction}: Explain some examples of methods in detail.\newline \textbf{Comment}: only one example of an actual technique is given here \\
\bottomrule
\end{tabular} 
}
\caption{Fine-grained error categories alongside example instruction and corresponding annotator comment pairs. We report the distribution classified by GPT-4 on interaction turns that are rated as \texttt{neutral} or \texttt{bad}.}\label{tab:error_analysis}
\end{center}
\end{table}

We find five major categories (Table \ref{tab:error_analysis}). The most common error type is \textbf{Unrequested change}, indicating that models fail to maintain answer consistency across turns. Models also fail to satisfy hard constraints, such as the location 
(``At the beginning of the answer...'') and length of the information (``Add one sentence about...'') specified in the instruction. This suggests that current models struggle to perform \textit{precise} actions, in agreement with findings from recent work \cite{Sun2023EvaluatingLL} on controlled generation. Finally, models struggle to coherently integrate new information into the existing answer text, often leading to imbalanced structure or awkward answer flow.

\paragraph{Retrieval analysis} The model could fail to follow the instruction due to retrieval failure instead of the language model. We randomly sample 20 turns with info-seeking instructions that are rated as \texttt{neutral} or \texttt{bad} and manually examine retrieval performance. We find that for 50\% of the turns, relevant passages are retrieved which contain sufficient information to follow the instruction, yet the language models fail to integrate the information into the previous answer. For 20\% of turns, some (but not all) relevant passages are retrieved. For the remaining 30\%, the retrieval system fails to retrieve any relevant information. This demonstrates room for improvement in \textit{both} components.

\section{Experiments: Automatically evaluating instruction following}\label{sec:exps}
Our analysis so far demonstrates that current LLMs often fail to follow users' instructions to revise long-form answers. Next, we examine whether they can serve as an \emph{evaluator} to assess whether a model-generated answer followed the instructions \cite{Dubois2023AlpacaFarmAS, Chiang2023CanLL, Zeng2023EvaluatingLL, Zheng2023JudgingLW, Liu2023BenchmarkingGA}.

\paragraph{Setting and data} Formally, the task is to predict the user rating $r_t = \mathbf{r}(q, i_{t}, a_{t-1}, a_{t})$, introduced in \S \ref{subsec:annot_process}. To simplify the task, we collapse \{\texttt{neutral}, \texttt{bad}\} into a single label. We randomly split the 1,260 turns into train/dev/test set, resulting in 888 training, 186 validation and 186 test instances.

\paragraph{Metrics and baselines} We evaluate using \textbf{Acc}uracy, \textbf{P}recision, \textbf{R}ecall and \textbf{F1} against the collected human labels $r_t$, with the \texttt{good} label as the positive class. We report two baselines: a \textbf{Random} baseline which randomly assigns a label according to the training data distribution, and a \textbf{Majority} baseline which always chooses the majority class in the training data.
We measure \textbf{Human} agreement by collecting a second set of human ratings for 65 randomly-sampled instructions, and measuring agreement with the original ratings.\footnote{Two authors and two annotators from the interaction annotation performed the annotation. We ensure the interaction annotators do not re-annotate their own previous interactions.}

\paragraph{Model} We experiment with \textbf{zero-shot} and \textbf{few-shot} prompting with \gptFourFull and a \textbf{fine-tuned} T5-large (770M) \cite{Raffel2019ExploringTL} model. For zero-shot prompting, we construct an instruction which specifies the criteria for the two ratings. Inspired by prior work on retrieving in-context examples \cite{rubin-etal-2022-learning}, we retrieve the $k$ turns from the training data whose instructions have the highest BM25 similarity with the test instruction. See Appendix \ref{subsec:edit_eval_auto} for details.

\begin{table}
\begin{center}
\small
\begin{tabular}{@{}lcccccc@{}}
\toprule
\textbf{Model} & \textbf{Acc} & \textbf{P} & \textbf{R} & \textbf{F1} & \textbf{\% g/b} \\
\midrule
{Majority} & 0.53 & 0.0 & 0.0. & 0.0 & 0/100 \\ 
{Random} & 0.47 & 0.42 & 0.31 & 0.36 & 38/62 \\
{\gptFourIcon zero-shot} & 0.63 & 0.57 & \textbf{0.91} & \textbf{0.70} & 75/25 \\ 
{\gptFourIcon one-shot} & 0.62 & 0.50 & 0.86 & {0.65} & 65/35 \\
{\gptFourIcon ten-shot} & \textbf{0.68} & \textbf{0.67} & 0.59 & 0.63 & 40/60 \\
T5 (finetuned) & 0.64 & 0.5 & 0.73 & 0.37 & 24/76 \\
\midrule 
\textbf{Human*} & 0.80 & 0.63 & 0.90 & 0.75 & 46/54 \\ 
\bottomrule
\end{tabular} 

\caption{Test set results on automatic evaluation for instruction following edits. We also report the \% of \textbf{g}ood and \textbf{b}ad ratings in the model predictions. }\label{tab:automatic_eval_test}
\end{center}
\end{table}

\paragraph{Results}
We report results on the test set in Table \ref{tab:automatic_eval_test}. GPT-4 zero-shot performs slightly worse than choosing the majority label; we found that it is biased heavily towards judging answers as \texttt{good}. Adding 10 in-context examples improves performance, while leaving a substantial gap relative to human agreement. Finetuned T5 only achieves similar accuracy with zero-shot \gptFourName. While recent work found that LLMs can rate responses as reliably as human~\cite{Chiang2023CanLL, Zheng2023JudgingLW}, our experiments show that they are not reliable for judging responses for specific instructions \cite{Zeng2023EvaluatingLL, Liu2023BenchmarkingGA}.

\section{Related Work}\label{sec:related_work}

\paragraph{Instruction following} A number of recent efforts have collected instruction following datasets to train and evaluate LLMs. Some datasets contain crowd-sourced interactions in the wild \cite{alpaca, dolly, Kopf2023OpenAssistantC, zhao2023wildchat, Zheng2023LMSYSChat1MAL}, while others target specific tasks such as summarization \cite{Liu2023BenchmarkingGA} and controllable generation \cite{Zhou2023ControlledTG}. Our work focuses specifically on instruction-following for writing assistance and text editing. A number of prior works have curated instruction-based editing benchmarks \cite{DwivediYu2022EditEvalAI, Shu2023RewriteLMAI, Raheja2023CoEdITTE, Zhang2023XATUAF}. However, these efforts consist of post-hoc synthetic instructions derived from previous datasets, consisting of a limited set of edit instructions. Our data consists of diverse instructions issued by expert users interacting with LLMs.

\paragraph{LLM-based Evaluation} Recent works have investigated using an LLM as an evaluator in place of human evaluation \cite{Chiang2023CanLL, Zheng2023JudgingLW, alpaca_eval} for instruction following. A line of work shows that LLMs can judge responses as reliable as human annotators \cite{Zheng2023JudgingLW, alpaca_eval}. Recent work \cite{Zeng2023EvaluatingLL} points out that evaluating open-ended instruction following exhibit large subjectivity and thus low human performance. The authors curate a set of instructions that are objective and craft adversarial response to stress test LLMs, finding that the models struggle at evaluating responses for such instructions while humans exhibit high agreement. Our study also shows that LLMs struggle at evaluating responses for precise instructions collected in a \textit{realistic} setting, complementing prior work.

\section{Conclusion}
We present \dsname, a dataset with expert-issued writing instructions to revise a long-form answer to a research question using relevant documents. Our analysis and experiments show that current LLMs (including GPT-4) cannot perform this task reliably yet and identify common failure patterns.
We are optimistic that \dsname will be a useful resource for several research directions. First, the model revisions that were judged as successful by humans, and answers edited manually by humans when models failed, can be used as training data to improve models' instruction following abilities. Second, human judgments of model revision quality can be used to develop more accurate reward models for writing assistance. Finally, the instructions in \dsname can be used as inputs to evaluate the performance of future models; either through human judgments, or by future reward models.

\section*{Acknowledgments}
We thank Nicholas Behrje, Sergey Feldman, Sebastian Joseph, Yoonjoo Lee, Clara Na, Matthias Rémy, Prasann Singhal, Zayne Sprague, Irina Temnikova, Albert Yu and other annotators for participation in creating the dataset. We thank the Semantic Scholar team and UT Austin NLP lab for helpful discussion and piloting the annotations. We thank Pradeep Dasigi, Sachin Kumar, and Leo Zeyu Liu for feedback to improve the paper draft.

\paragraph{Author Contributions:} Fangyuan Xu led the project and performed all the technical contributions including designing and building the annotation interface, dataset collection, running the experiments and performing the data analysis. Fangyuan also contributed to project scoping and ideation and led the writing of the paper. Bailey Kuehl helped with collecting the instruction data, including piloting the task, giving feedback and hiring and managing the Upwork annotators. David Wadden, Kyle Lo and Luca Soldaini were mentors of the project during and after Fangyuan's internship, contributed to project scoping, experiments design, ideation and direction throughout the project. David Wadden and Kyle Lo contributed to paper writing and Luca Soldaini contributed to annotation interface development. Eunsol Choi provided mentorship throughout the project, including project scoping, ideation, experiement design and paper writing.

\section*{Ethics Statement}
Our annotation protocol has been determined to be exempted from review by an IRB board at UT Austin. All data collected will be made publicly available under the MIT license. The data collection process did not require any information that can be used to uniquely identify individual annotators. We manually examine the annotation and make sure no such information or offensive content is present in the data collected.

\section*{Limitations}
Our study focus on the domain of writing answers for scientific questions and more specifically, NLP research questions. While we believe some of our findings could generalize to other domains such as creative writing, our data might only represent a subset of instructions that occur in other writing tasks or domain. We only cover the English language, and future work might collect similar data for other languages or look into leveraging our data for cross-lingual transfer.




\bibliography{anthology,custom}
\bibliographystyle{acl_natbib}

\appendix

\section{Appendix}
\label{sec:appendix}

\subsection{Implementation details for the interaction system}\label{sec:system_details} 

\begin{figure*}[t!]
 \vspace{-2em}
    \centering
    \includegraphics[trim={0 0 0 0},clip,width=\textwidth]{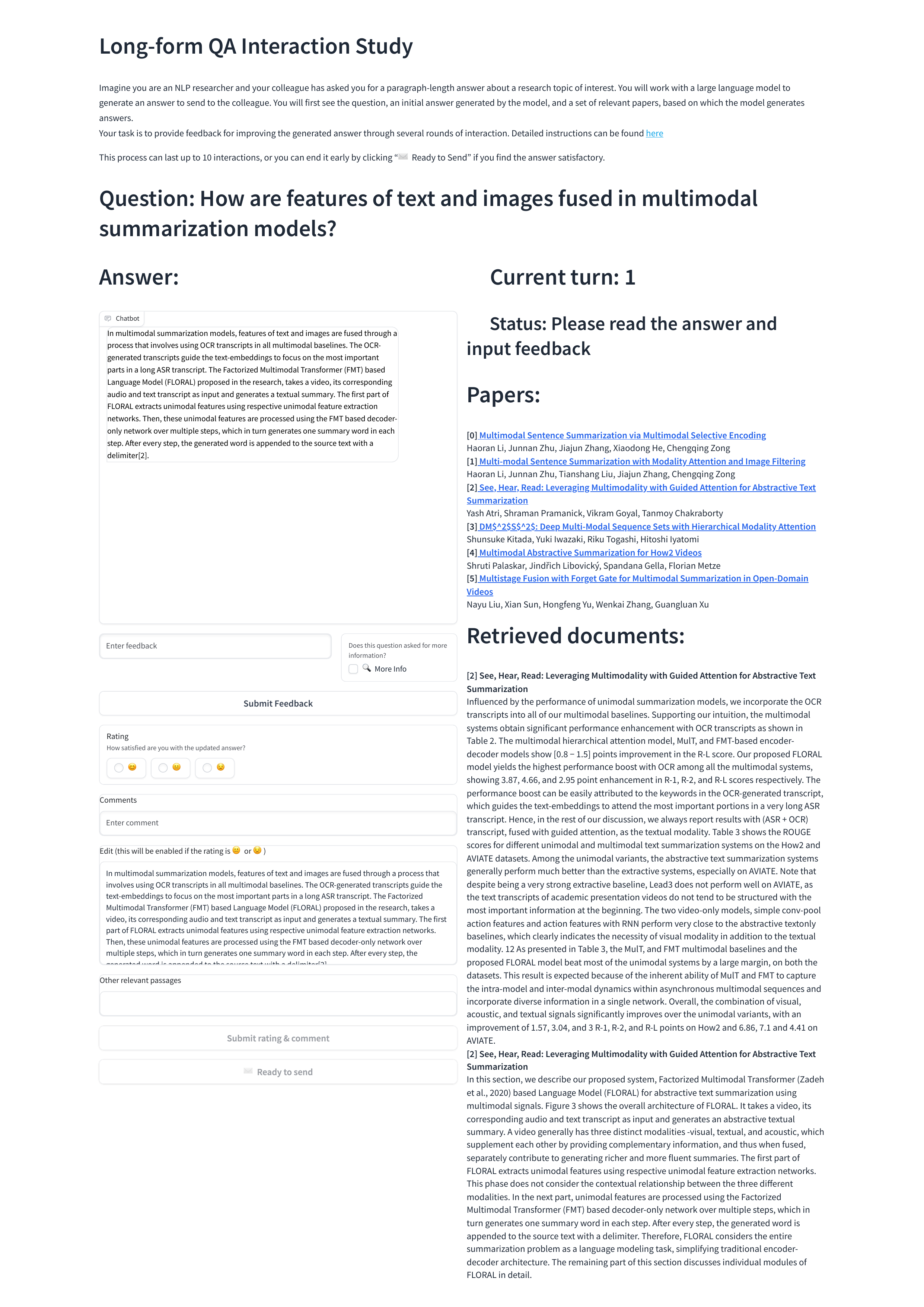}
    \caption{A screenshot of our annotation interface.}
    \label{fig:annotation_interface}
\end{figure*}

We implement the annotation interface with gradio\footnote{\url{https://www.gradio.app/}}. Figure \ref{fig:annotation_interface} presents a screenshot of our annotation interface. We use the OpenAI API\footnote{\url{https://platform.openai.com/docs/api-reference}} for \texttt{gpt}-family model inference and the Together AI API \footnote{\url{https://docs.together.ai/reference/inference}} for \texttt{llama2}. Below are the prompt templates we use.

\paragraph{Prompt for generating initial answer} We include the top 5 retrieved passages from the document set to generate the initial answer. \texttt{title\_i} and \texttt{passage\_i} refers to the title and text of the i-th retrieved passages using \texttt{question}  as the query.

\begin{quote}
\textbf{SYSTEM MESSAGE:} You are a helpful assistant which answers a question based on a given set of documents. Please add reference (e.g. [0]) to the document in the answer. \\

\textbf{USER MESSAGE:} Documents:

[0] Paper title: \texttt{title\_0}

Passage: \texttt{passage\_0}

[1] Paper title: \texttt{title\_1}

Passage: \texttt{passage\_1}

[2] Paper title: \texttt{title\_2}

Passage: \texttt{passage\_2}

[3] Paper title: \texttt{title\_3}

Passage: \texttt{passage\_3}

[4] Paper title: \texttt{title\_4}

Passage: \texttt{passage\_4}

Question: \texttt{question} \\
Answer:
\end{quote}

\paragraph{Prompt for answer revision of information seeking instructions} We first prompt the model to generate a summary $s_{t}$ given the previous answer ($a_{t-1}$), the retrieved passages and the instruction $i_{t}$. 

\begin{quote}
\textbf{SYSTEM MESSAGE:} You are a helpful assistant which generates an intermediate answer given a feedback to the previous answer. Please add reference (e.g. [0]) to the document in the answer.\\

\textbf{USER MESSAGE:} Documents:

[0] Paper title: \texttt{title\_0}

Passage: \texttt{passage\_0}

[1] Paper title: \texttt{title\_1}

Passage: \texttt{passage\_1}

[2] Paper title: \texttt{title\_2}

Passage: \texttt{passage\_2}

[3] Paper title: \texttt{title\_3}

Passage: \texttt{passage\_3}

[4] Paper title: \texttt{title\_4}

Passage: \texttt{passage\_4}

Previous answer:$a_{t-1}$

Feedback:$i_{t}$

Generate an answer to the feedback.
\end{quote}

We then prompt the model to generate an updated answer, given the original question, the answer in the previous turn $a_{t-1}$, the instruction $i_{t}$ and the summary $s_{t}$. For \llamachatName, we add an additional instruction at the end ``Do not generate “Sure”, directly generate the updated answer. Updated answer: ''.

\begin{quote}
\textbf{SYSTEM MESSAGE:} You are a helpful assistant which generates an intermediate answer given a feedback to the previous answer. Please add reference (e.g. [0]) to the document in the answer.\\

\textbf{USER MESSAGE:} 
Original question: \texttt{question} 

Previous answer: $a_{t-1}$ 

Feedback: $i_{t}$

Extra information related to the feedback: $s_{t}$

Answer the original question and incorporate the feedback to the previous answer. Preserve all relevant information from previous answer.
\end{quote}

\paragraph{Prompt for answer revision of stylistic instructions} We include the original question, answer in the previous turn ($a_{t-1}$) and the instruction $i_{t}$ in the prompt. For \llamachatName, we add an additional instruction at the end ``Do not generate “Sure”, directly generate the updated answer. Updated answer: ''.

\begin{quote}
\textbf{SYSTEM MESSAGE:} You are a helpful assistant which answers a question based on a given set of documents. Please add reference (e.g. [0]) to the document in the answer. \\

\textbf{USER MESSAGE:} Original question: \texttt{question}

Previous answer: $a_{t-1}$

Feedback: $i_{t}$

Answer the original question and incorporate the feedback to the previous answer. Preserve all relevant information from previous answer.
\end{quote}

\paragraph{Prompt length} On average, the prompt for generating an answer revision for information seeking instructions (with question, previous answer, instruction and retrieved passages) consists of 607 words. The retrieved passages consists of 560 words on average. The prompt for answer revision for stylistic instructions (with question, previous answer and instruction) consists of 458 words on average.

\subsection{Examples}\label{subsec:data_examples}
Examples turns with \texttt{good}, \texttt{neutral} and \texttt{bad} rating can be found in Table \ref{tab:rating_examples}. Example of questions annotated (\S \ref{subsec:question_annotation}) can be found in Table \ref{tab:example_questions}. 

\subsection{Implementation details for automatic instruction analysis with GPT-4}\label{sec:instruction_analysis_auto}

\paragraph{Manual analysis} We randomly sample 50 interaction turns and manually group the instructions into fine-grained categories. We discover 8 fine-grained categories: requesting for more information (46\%); expanding answer coverage (6\%); asking for clarification (6\%), categorization (6\%), removal (10\%), specific edits (2\%), condense (12\%) and reorganization (10\%).

\paragraph{GPT-4 analysis} We construct a prompt with definition and in-context examples for each of the category. The prompt we use to perform automatic instruction analysis in in Table \ref{tab:instruction_analysis_prompt}. We prompt \texttt{gpt-4} with temperature of 0 and top\_p=1. The distribution classified by GPT-4 is presented in \ref{tab:example_instructions}. We see a slight distribution differences between the large-scale GPT-4 analysis and small-scale manual analysis, with GPT-4 identifying more instructions as ``specific edits'' and less as ``categorization''. We merge categories with fewer than 5\% of examples into ``Others''.

\paragraph{More statistics} We report additional statistics for each instruction type in Table \ref{tab:edit_stats}. We filter out turns where the response is rated as \texttt{neutral} or \texttt{bad} with an edited answer, resulting in a total of 815 turns. Among the coarse-grained instruction types, we see that stylistic instructions are longer, yet with less edit ratio compared to information seeking ones. Specific edit instructions are the longest, while instructions asking for more information requires the most amount of edits to the previous answer (highest edit ratio).


\begin{table*}
\begin{center}
\footnotesize
\begin{tabular}{@{}lcccccc@{}}
\toprule
\multirow{2}{*}{\textbf{Type}} & \multicolumn{3}{c}{\textbf{\# words}} & \multirow{2}{*}{\textbf{Len ratio}} & \multirow{2}{*}{\textbf{Edit distance}} & \multirow{2}{*}{\textbf{Edit ratio}}\\
& inst & src & target & \\ 
\midrule
More information & 17 & 227 & 340 & \textbf{1.60} & 162 & \textbf{0.85} \\
Expand coverage & 39 & 295 & 398 & 1.48 & 195 & 0.78 \\ 
Clarification & 27 & 287 & 324 & 1.23 & 137 & 0.58 \\ 
Remove & 24 & 396 & 339 & 0.87 & 91.31 & 0.24 \\
Specific edit & \textbf{75} & 278 & 325 & 1.22 & 93 & 0.38 \\ 
Condense & 15 & 388 & 234 & 0.64 & 213 & 0.55 \\ 
Reorganize & 26 & 391 & 339 & 0.87 & 91.31 & 0.24 \\ 
Others & 23 & 327 & 342 & 1.36 & 212 & 0.88 \\ 
\midrule
Stylistic & \textbf{29} & 361 & 316 & 0.94 & 141 & 0.42  \\
Info seeking & 22 & 259 & 355 & \textbf{1.60} & 176 & \textbf{0.87} \\ 
\midrule 
GPT-4 & 24 & 260 & 286 & 1.28 & 145 & \textbf{0.70} \\ 
GPT-3.5 & 23 & 289 & 332 & \textbf{1.34} & 145 & 0.68 \\ 
Llama & \textbf{29} & 379 & 387 & 1.20 & 173 & 0.56 \\ 
\midrule
\textbf{Total} & 26 & 310 & 336 & 1.27 & 158 & 0.64 \\
\bottomrule
\end{tabular} 
\caption{We report statistics for each individual turns in our dataset: number of words in the instruction, number of words in the previous answer (src); number of words in the answer (target); the average length fraction between the source and the target; the token-level edit distance; and the fraction between th edit distance and the source text. We filter out turns where the response is rated as \texttt{neutral} or \texttt{bad} and there isn't an edited answer.}\label{tab:edit_stats}
\end{center}
\end{table*}

\subsection{Implementation details for error analysis}\label{sec:comment_analysis_auto}

\paragraph{Manual analysis} We randomly sample 10 interaction turns per model which are rated as \texttt{neutral} or \texttt{bad}, resulting in 30 total turns. We then group them into 5 fine-grained categories based on the question and comment. 

\paragraph{GPT-4 analysis} The prompt we use to perform automatic instruction analysis in in Table \ref{tab:comment_analysis_prompt}. We prompt \texttt{gpt-4} with temperature of 0 and top\_p=1.

\subsection{Implementation details for automatic evaluation of instruction following}\label{subsec:edit_eval_auto}

\subsubsection{GPT-4} We prompt \texttt{gpt-4-1106-preview} \footnote{\url{https://platform.openai.com/docs/models}} for this task, whose context window can fit the ten-shot examples. We decode with temperature of 0 and top\_p=1. Below are the prompt templates for the different baselines.

\paragraph{Zero-shot prompt template} The model is given an instruction, the question, the answer in the previous turn $a_{t-1}$, the instruction $i_{t}$ and the answer in the current turn $a_{t}$.
\begin{quote}
\textbf{SYSTEM MESSAGE:} You are a helpful assistant in evaluating the quality of the outputs for a given instruction to update an answer for an question. Your goal is to score a given updated answer for the given instruction.

\textbf{USER MESSAGE:} Score the updated answer for the given instruction by comparing it with the original answer.You should give one of the two ratings: good, or bad.

Give a bad rating if the updated answer either (1) only partially followed the instruction (for example, the instruction specify an edit of a certain length or at a certain location but the updated answer didn't follow the constraint)  or (2) introduced changes that are not mentioned in the instruction compared to the original answer or (3) provided a vague answer (e.g. without naming actual method) or (4) added the requested information but made the answer less coherent/correct.

Only give a good rating if the answer COMPLETELY followed the instruction and didn't have ANY issues mentioned above.
Your response should be ONLY the ratings.

\# Question:

\# \texttt{question}
\\

\# Original answer: 

\# $a_{t-1}$
\\

\# Instruction:

\# $i_{t}$
\\

\# Updated answer:

\# $a_{t}$ 
\\

\# Rating of the Updated answer:

\#
\end{quote}

\paragraph{Few-shot prompt template} Each few-shot example $j$ contains the question, the answer in the previous turn, the instruction, the answer in the current turn and the rating of the answer. 

\begin{quote}
\textbf{SYSTEM MESSAGE:} You are a helpful assistant in evaluating the quality of the outputs for a given instruction to update an answer for an question. Your goal is to score a given updated answer for the given instruction.

\textbf{USER MESSAGE:} Score the updated answer for the given instruction by comparing it with the original answer.You should give one of the two ratings: good, or bad.

Give a bad rating if the updated answer either (1) only partially followed the instruction (for example, the instruction specify an edit of a certain length or at a certain location but the updated answer didn't follow the constraint)  or (2) introduced changes that are not mentioned in the instruction compared to the original answer or (3) provided a vague answer (e.g. without naming actual method) or (4) added the requested information but made the answer less coherent/correct.

Only give a good rating if the answer COMPLETELY followed the instruction and didn't have ANY issues mentioned above.
Your response should be ONLY the ratings.

\# Question:

\# \texttt{question\_j}
\\

\# Original answer: 

\# $a_{(t-1)j}$
\\

\# Instruction:

\# $i_{tj}$
\\

\# Updated answer:

\# $a_{tj}$ 
\\

\# Rating of the Updated answer:

\# $r_{tj}$
\\

\# Question:

\# \texttt{question}
\\

\# Original answer: 

\# $a_{t-1}$
\\

\# Instruction:

\# $i_{t}$
\\

\# Updated answer:

\# $a_{t}$ 
\\

\# Rating of the Updated answer:

\#
\end{quote}

\subsubsection{T5}
We finetune the model to output target sequence ``Rating : $r_{i}$'' with input sequence ``Original question: $q$ Previous answer: $a_{t-1}$ Instruction: $i_{t}$ Updated answer: $a_{t}$'', where $r_{i}$ is the traget binary rating, $q$ is the question, $a_{t-1}$ is the answer in the previous turn, $i_{t}$ is the instruction and $a_{t}$ is the answer in the current turn.

We use a batch size of 16 and an initial learning rate of \texttt{1e-4} with Adam optimizer and a linear learning rate schedule. We train the model for 5 epochs and choose the checkpoint with the best validation accuracy. The hyperparameters are manually searched by the authors.


\subsection{Analysis on question generation with GPT-4}\label{sec:question_generation}
During the initial phase of the project, we explored an automatic version of the pipeline described in \S \ref{subsec:question_annotation} by using GPT-4 to generate questions. We describe the pipelines below:

\paragraph{Step 1: Extracting and filtering related works} We extract related work paragraphs from papers published in NLP venues (ACL, EMNLP, NAACL, etc.) using S2ORC \cite{lo-etal-2020-s2orc}. We extract the papers cited and filter out paragraphs with more than 30\% of the papers missing from S2ORC or with less than 4 cited papers available. This gives us an initial set of ($q$, \textbf{D}).

\paragraph{Step 2: Prompting LMs to generate questions} For each related work paragraph, we prompt GPT-4 to generate five questions using the below prompt. We set the temperature to be 1 and top\_p=1. \texttt{paper\_title} is the title of the paper from which the related work paragraph is extracted.

\begin{quote}
\textbf{SYSTEM MESSAGE:} You are a helpful assistant which generates five questions that the paragraph is addressing. The question should require multiple sentences to answer. Don't ask multiple sub questions in a single question.  Don't refer to specific paper in the question. \\
\textbf{USER MESSAGE:} Paper title: \texttt{paper\_title} \\ 
Passage: \texttt{related work paragraph} \\ 
\end{quote}

\paragraph{Step 3: Question filtering} We employ two filtering steps to filter out questions that do not fulfill the desiterata that we described in \S \ref{subsec:question_annotation}. First, we use a heuristic rule to filter out questions that contain keywords which make the question anchored to specific papers (e.g. ``this paper'', ``mentioned'', ``author''). This process filters out 23\% of the questions generated.

We then prompt GPT-4 on whether the generated question (\texttt{question}) can be answered by the related work paragraph using the following prompt. This process filters out 12\% of the question which are rated as \texttt{No} or \texttt{Paratially} by GPT-4.

\begin{quote}
\textbf{SYSTEM MESSAGE:} You are a helpful assistant. \\
\textbf{USER MESSAGE:} Check if the paragraph answers the question. Reply Yes, No or Partially.\\

Question: \texttt{question} 

Paragraph: \texttt{related work paragraph} \\ 
\end{quote}

\paragraph{Manual examination} We conduct a manual examination on the quality of the $(q, \mathbf{D)}$ pairs generated through this automatic pipeline. We look at both the question and an initial answer generated by GPT-4 given $q, \mathbf{D}$, using the prompt described in \S \ref{sec:system_details}. Two of the authors randomly sample 50 (question, initial answer) pairs to examine both the quality of the question and the initial answer.

We found that only 60\% of the questions fulfill our desiderata described in \S \ref{subsec:question_annotation}. 28\% of the question do not make sense to a researcher (e.g. ``How do previous approaches to multi-task learning and domain adaptation try to create a universal representation space using encoders and architectures?''), 6\% of the questions are too specific or niche (e.g. ``What are the primary components used by the VGVAE model to represent the semantics and syntax of a sentence in a monolingual setting?'') and 6\% of the questions do not stand alone (e.g. ``How do labels like "other" or "information providing" act indicate issues with these models?''). For the questions that fulfill our desiderata, GPT-4 is unable to generate a good initial answer for 26\% of them, often due to retrieval failure (e.g. ``The documents provided do not discuss the potential issues with using word segmentation information and dependency trees for Chinese Named Entity Recognition (NER) as extra features.''). We note that this is not necessarily an issue with the retrieval system, but also an indication that the question could be irrelevant to the papers cited.

Overall, we find that this pipeline does not produce high quality ($q$, $D$) pairs needed for \dsname. Hence we proceed with human annotations as described in \S \ref{subsec:question_annotation}.

\begin{table}
\footnotesize
\begin{center}
\begin{tabular}{p{7.5cm}}
\toprule
How are features of text and images fused in multimodal summarization models? \\
What are the different methods proposed for improving factual consistency in summaries generated by summarization systems? \\
Is it possible to extract the data used to train a language model and if so, how to do it? \\
How are pre-training corpora constructed for language models? \\
How are cross-lingual language models pre-trained? \\
\bottomrule
\end{tabular}
\caption{Example questions annotated (\S \ref{subsec:question_annotation}).}
\label{tab:example_questions}
\end{center}
\end{table}

\begin{figure}
\begin{framed}
\small
\textbf{Require:} Language model $M$, Retrieval model $R$, User $U$ \\
\textbf{Input:} question $q$ and a set of relevant papers $D$ \\
\textbf{Output:} a series of interaction $\mathcal{I}$, each consists of (instruction $i$, answer $a$, rating $r$, comment $c$, edited answer $a'$)\\
\begin{algorithmic}[1]
\STATE $\mathcal{I} \gets \emptyset$ \\
$a_{0} = M(q, R(q, D))$ \COMMENT{\textcolor{blue}{\textit{Generate an initial answer}}} \\ 
$a_{p} \gets a_{0}$ \COMMENT{\textcolor{blue}{\textit{Set initial answer as previous answer}}}
\FOR{$t  \in \{1, \dots, 10 \}$}
     \STATE $i_{t}, I_{t} = U(q, a_{0}, ... a_{t-1}, D)$   \COMMENT{\textcolor{blue}{\textit{User issues an instruction $i_{t}$, and $I_{t}$, the type of the instruction.}}} \\
     \STATE $p_{t} \gets \emptyset$
      \IF{$ I_{t} == \texttt{info} $}
          \STATE $p_{t} =  R(i_{t}, D)$
     \ENDIF
     \STATE $a_{t} = M(q, a_{p}, p_{t}, i_{t})$  
     \STATE $r_{t}, c_{t} = U(q, i_{t}, prev\_ans, a_{t})$  \COMMENT{\textcolor{blue}{\textit{User provides a rating and comment for the updated answer.}}} 
     
     \STATE $a_{t}' \gets a_{t}$
     \IF{$ r_{t} \in \{\texttt{neutral}, \texttt{bad}\} $}
          \STATE $a_{t}' =  U(q, i_{t}, a_{t-1})$  \COMMENT{\textcolor{blue}{\textit{User optionally edits the updated answer.}}}
     \ENDIF
     \STATE $\mathcal{I} \gets \mathcal{I} \cup \{(i_{t}, a_{t}, r_{t}, c_{t}, a_{t}')\}$
     \IF{$ U(q, a_{t}') == \textbf{Good} $}
          \STATE{break} \COMMENT{\textcolor{blue}{\textit{Interaction ends if the user is satisfy with the answer at this turn.}}}
     \ELSIF{ $r_{t} != \textbf{Bad } \OR$ $a_{t} != a_{t}'$}
          \STATE $a_{p} \gets a_{t}'$ \COMMENT{\textcolor{blue}{\textit{Update previous answer if answer in this turn is not rated as bad \textbf{OR} edited by user.}}}
     \ENDIF
\ENDFOR
\end{algorithmic}
\end{framed}
\caption{We present the workflow for collecting \dsname. Detailed description for each step is in \S \ref{sec:task_setup}.}\label{alg:interaction_procedure}
\end{figure}





\begin{table*}[]
\scriptsize
\resizebox{1\textwidth}{!}{
\begin{tabular}{@{}p{1cm}p{16cm}@{}}
\toprule
\textbf{Rating} & \textbf{Example} \\
\midrule
\texttt{good} & \textbf{Original question:} Are there any datasets available fro table-to-text generation tasks in NLP? \newline
\textbf{Previous answer}: Yes, there are datasets available for table-to-text generation tasks in NLP. State-of-the-art models for data-to-text generation tasks have been trained and evaluated on various datasets. One commonly used dataset is the WebNLG dataset [0]. The DART model, for example, has achieved new state-of-the-art results on all WebNLG 2017 test set splits by fine-tuning T5-large on DART [0]. Other competitive models, both pretrained and not pretrained, have also been evaluated on this dataset [0] \newline 
The DART dataset has been used for data augmentation and has shown to improve the performance of existing models on the WebNLG 2017 dataset [0].\newline The TOTTO dataset is another dataset used for table-to-text generation tasks [5]. However, state-of-the-art neural models have been found to struggle in generating faithful results on the TOTTO dataset, despite the high quality of the training data [5]. The performance gap between the models and the oracle upper-bound, which treats one of the references as a candidate, has been shown to be considerable [5].\newline 
The WikiTableT dataset focuses on generating Wikipedia article sections based on structured data [4]. However, the performance of state-of-the-art models on this dataset has not been explicitly mentioned in the provided documents.\newline 
Overall, while state-of-the-art models have achieved impressive results on these datasets, there is still room for improvement and further research in the field of data-to-text generation [0] [4] [5]. \newline 
\textbf{Instruction:} What metric was used to measure performance on these datasets? \newline
\textbf{Updated answer:} Yes, there are datasets available for table-to-text generation tasks in NLP. State-of-the-art models for data-to-text generation tasks have been trained and evaluated on various datasets. One commonly used dataset is the WebNLG dataset [0]. The participating systems on the WebNLG dataset were evaluated using three automatic metrics [1]. The DART model, for example, has achieved new state-of-the-art results on all WebNLG 2017 test set splits by fine-tuning T5-large on DART [0]. Other competitive models, both pretrained and not pretrained, have also been evaluated on this dataset [0].\newline 
The DART dataset has been used for data augmentation and has shown to improve the performance of existing models on the WebNLG 2017 dataset [0].\newline 
The TOTTO dataset is another dataset used for table-to-text generation tasks [5]. The evaluation on the TOTTO dataset used the BLEU metric, which measures n-gram overlap between the reference and the prediction at the corpus level [5]. However, state-of-the-art neural models have been found to struggle in generating faithful results on the TOTTO dataset, despite the high quality of the training data [5]. The performance gap between the models and the oracle upper-bound, which treats one of the references as a candidate, has been shown to be considerable [5].\newline 
The WikiTableT dataset focuses on generating Wikipedia article sections based on structured data [4]. The performance of state-of-the-art models on this dataset has not been explicitly mentioned in the provided documents.\newline 
Overall, while state-of-the-art models have achieved impressive results on these datasets, their performance was measured using various metrics such as BLEU and REP [1] [4] [5]. These metrics provide quantitative measures to assess the performance of models on these datasets and compare different approaches. However, there is still room for improvement and further research in the field of data-to-text generation [0] [4] [5]. \newline 
\textbf{Free-form comment:} The model provided the requested metrics. \\ 
\midrule
\texttt{neutral} & \textbf{Original question:}  What approaches have been proposed to edit a generated text to be factually consistent with a collection of evidence? \newline 
\textbf{Previous answer:} In the context of factuality correction, a “collection of evidence” refers to a set of information or data that supports or refutes a claim or statement. This collection typically consists of retrieved snippets or passages from various sources, such as web pages or documents, which are used to assess the factual consistency of the generated text and make necessary revisions [1].\newline 
Approaches that have been proposed to edit a generated text to be factually consistent with a collection of evidence include:\newline  1. Fact extraction or attention on fact triples: This approach involves extracting and incorporating factual information from the evidence collection into the generated text. Fact triples, which consist of subject-predicate-object statements, can be used to represent these extracted facts [5].\newline 
2. Natural language inference or question-answering models for consistency checking: These models can be employed to verify the factual consistency of the generated text with the evidence collection. By comparing the generated text with the evidence, these models can identify any inconsistencies and suggest appropriate modifications [5].\newline 
3. Training the model on artificial datasets: In order to improve factuality correction, the model can be trained on artificial datasets that contain both generated text and corresponding evidence. This approach helps the model learn to align the generated text with the evidence and make necessary revisions [5].\newline 
4. Adversarial data generation process using Infilling Language Models: This approach involves generating adversarial examples by filling in missing information in the generated text. By iteratively generating and updating the text, the model can be fine-tuned to ensure factual consistency with the evidence [4].\newline 
5. Factual error correction by editing inconsistent parts in generated summaries: This approach focuses on identifying and editing specific parts of the generated text that are inconsistent with the evidence. By revising these inconsistent parts, the overall factuality of the text can be improved [5].\newline 
6. Factual error correction by rewriting claims to better align with retrieved evidence: This approach involves rewriting or modifying the claims made in the generated text to align them more closely with the evidence. By making these adjustments, the generated text can be made more factually consistent [6].\newline 
These approaches from papers [0], [2], [3], [4], [5], and [6] aim to enhance the factuality of generated texts by incorporating evidence and making necessary changes to ensure factual consistency. \newline 
\textbf{Instruction:} Now keep the content and the order of the text as it is, but re-write the text into 1 paragraph only \newline 
\textbf{Updated answer}: Approaches that have been proposed to edit a generated text to be factually consistent with a collection of evidence include: extracting and incorporating factual information from the evidence collection into the generated text using fact extraction or attention on fact triples [5]; employing natural language inference or question-answering models to verify the factual consistency and suggest modifications [5]; training the model on artificial datasets that contain both generated text and corresponding evidence to align the text with the evidence [5]; generating adversarial examples by filling in missing information in the text and fine-tuning the model for factual consistency [4]; identifying and editing specific parts of the text that are inconsistent with the evidence to improve factuality [5]; rewriting or modifying the claims made in the text to align them more closely with the evidence [6]. These approaches aim to enhance the factuality of generated texts by incorporating evidence and making necessary changes to ensure factual consistency [0] [2] [3] [4] [5] [6]. [1] \newline
\textbf{Free-form comment:} It removed a lot from the text, including the definition \\ 
\bottomrule
\end{tabular} 
}
\caption{Example turns rated as \texttt{good}, \texttt{neutral}. The \texttt{bad} example is in Table \ref{tab:rating_examples_2}.}
\label{tab:rating_examples}
\end{table*}

\begin{table*}[]
\scriptsize
\resizebox{1\textwidth}{!}{
\begin{tabular}{@{}p{1cm}p{16cm}@{}}
\toprule
\textbf{Rating} & \textbf{Example} \\
\midrule
\texttt{bad} & \textbf{Original question:} Is it possible to extract the data used to train a language model and if so, how to do it? \newline 
\textbf{Previous answer:} Yes, it is possible to extract the data used to train a language model through extraction attacks. These attacks involve recovering specific training examples from a language model by leveraging its output [0]. The goal is to reconstruct verbatim training examples rather than just representative or "fuzzy" examples [0].\newline 
The process of performing extraction attacks typically involves several steps. First, the attacker generates probing queries designed to elicit specific responses from the language model [0]. These queries are carefully crafted to contain clues about the training data. Then, the attacker analyzes the responses generated by the language model to identify potential matches with the training data [0]. By observing patterns and similarities in the model\'s output, the attacker can infer the presence of specific training examples.\newline 
The attacker iteratively refines their probing queries based on the model responses, gradually narrowing down the potential training examples [0]. This iterative refinement process involves generating new queries that are tailored to elicit more precise responses from the language model. Once the attacker has identified potential training examples, they can reconstruct the verbatim content by combining multiple queries and responses [0]. By carefully analyzing the model\'s behavior, the attacker can piece together the training data that the language model has memorized.\newline 
Extraction attacks have been shown to be practical and efficient, allowing adversaries to extract hundreds of training examples from a language model [0][5][7]. These attacks have been successfully performed against various language models, including GPT-2 and clinical language models like ClinicalBERT [5][7]. Furthermore, extraction attacks can succeed even when a specific training example appears only a few times in the data [0].\newline  
In addition to extraction attacks, membership inference attacks can also be used to extract training data from language models. These attacks focus on determining whether a given example is part of the training data [5]. Adversaries train shadow models that mimic the behavior of the attacked model and use them to train an auditing model that can predict membership in the training dataset [5]. Membership inference attacks have been performed on different NLP models, including masked language models like GPT-2 and clinical language models such as ClinicalBERT, successfully extracting training data from these models [5][7].\newline 
Therefore, both extraction attacks and membership inference attacks can be used to extract the data used to train a language model [0][5][7]. \newline 
\textbf{Instruction:} Combine the second and third paragraphs into one paragraph of no more than five sentences in which the steps to perform an extraction attack are listed. \newline 
\textbf{Upadted answer:} Yes, it is possible to extract the data used to train a language model through extraction attacks. These attacks involve recovering specific training examples from a language model by leveraging its output [0]. The process of performing extraction attacks typically involves several steps.\newline 
First, the attacker generates probing queries designed to elicit specific responses from the language model [0]. These queries are carefully crafted to contain clues about the training data. Then, the attacker analyzes the responses generated by the language model to identify potential matches with the training data [0]. By observing patterns and similarities in the model's output, the attacker can infer the presence of specific training examples.\newline 
The attacker iteratively refines their probing queries based on the model responses, gradually narrowing down the potential training examples [0]. This iterative refinement process involves generating new queries that are tailored to elicit more precise responses from the language model. Once the attacker has identified potential training examples, they can reconstruct the verbatim content by combining multiple queries and responses [0]. By carefully analyzing the model's behavior, the attacker can piece together the training data that the language model has memorized.\newline 
Extraction attacks have been shown to be practical and efficient, allowing adversaries to extract hundreds of training examples from a language model [0][5][7]. These attacks have been successfully performed against various language models, including GPT-2 and clinical language models like ClinicalBERT [5][7]. Furthermore, extraction attacks can succeed even when a specific training example appears only a few times in the data [0].\newline 
In addition to extraction attacks, membership inference attacks can also be used to extract training data from language models. These attacks focus on determining whether a given example is part of the training data [5]. Adversaries train shadow models that mimic the behavior of the attacked model and use them to train an auditing model that can predict membership in the training dataset [5]. Membership inference attacks have been performed on different NLP models, including masked language models like GPT-2 and clinical language models such as ClinicalBERT, successfully extracting training data from these models [5][7].\newline 
Therefore, both extraction attacks and membership inference attacks can be used to extract the data used to train a language model [0][5][7]. [0] \newline 
\textbf{Free-form comment:} The model did not shorten the answer as requested. \\
\bottomrule
\end{tabular} 
}
\caption{Example turns rated as \texttt{bad} by the annotator.}
\label{tab:rating_examples_2}
\end{table*}

\begin{table*}[]
\scriptsize
\resizebox{\textwidth}{!}{%
\begin{tabular}{@{}p{15cm}@{}}
\toprule

You are given an instruction to improve an answer to a question. Your job is to assign the instruction into one of the category below.
\newline
\newline
Example 1 \newline
\# Original question: Are there any datasets available fro table-to-text generation tasks in NLP?
\# Instruction: How well do state-of-the-art models perform on these datasets? \newline
\# Category: asking for more information \newline
\# Explanation: This category of instruction ask for more information related to the question (e.g. example, limitation, evaluation results, background information, citations, definition). \newline
\newline
Example 2\newline
\# Original question: How has prior work study and evaluate robustness of multimodal models?\newline
\# Instruction: As first sentence of the text, add a very short one sentence definition of what are "multimodal models"\newline
\# Category: asking for more information \newline
\# Explanation: This category of instruction ask for more information related to the question (e.g. example, limitation, evaluation results, background information, citations, definition).\newline
\newline
Example 3\newline
\# Original question: How did prior work explore learning from human feedback for different NLP tasks?\newline
\# Instruction: Try to add very short information also from papers 0,1,3,5,6,7,8. Keep the rest of the text exactly as it is.\newline
\# Category: expand answer coverage\newline
\# Explanation: This category of instruction asks for more information directly answering the question, usually from multiple papers.\newline
\newline
Example 4\newline
\# Original question: What are the existing approaches for the lip-reading task?\newline
\# Instruction: You mention that some methods use a network to predict phoneme probabilities and were trained using CTC loss and that it was trained using YouTube videos. Does this mean that such methods require phoneme  level labels for the data to be trained, or does word level transcriptions work?\newline
\# Category: asking for clarification\newline
\# Explanation: This category of instruction asks for clarification of a concept already in the answer.\newline
\newline
Example 5\newline
\# Original question: What are the prior efforts that develop RL environments with language-informed tasks?\newline
\# Instruction: Organize the papers better in categories of what aspect of language-conditioned RL tasks they are tackling.\newline
\# Category: categorization\newline
\# Explanation: This category of instruction requests to categorize the content in the answer.\newline
\newline
Example 6\newline
\# Original question: How can information stored in knowledge bases be injected into language models task-specific model fine-tuning?\newline
\# Instruction: Please shorten the answer to a maximum of 6 paragraphs.\newline
\# Category: condense\newline
\# Explanation: This category of instruction requests to shorten/summarize the answer.\newline
\newline
Example 7\newline
\# Original question: How can I quantize the weights of a generative language model and still achieve reasonable model performance?\newline
\# Instruction: Without changing anything else, delete every instance of the string "(Feedback)", and delete the string "[1] “Compression of Generative Pre-trained Language Models via Quantization”" at the end of the answer.\newline
\# Category: remove\newline
\# Explanation: This category of instruction requests to remove/delete certain part of the answer.\newline
\newline
Example 8\newline
\# Original question: What techniques have been proposed to measure and evaluate the effectiveness of task-oriented dialogue systems?\newline
\# Instruction: Move the paragraph starting with "Finally, there are also various task-specific evaluation metrics" to after the discussion of user satisfaction modeling, and make sure to cite paper [2] in that paragraph.\newline
\# Category: reorganize\newline
\# Explanation: This category of instruction requests to reorganize the answer(e.g. move things around, etc.).\newline

Example 9\newline
\# Original question: What methods have been proposed to categorize the kinds of editing operations that occur during text simplification?\newline
\# Instruction: Immediately after this sentence: "In the process of text simplification, different text-editing operations are used. " add this sentence "The editing operations that occur during text simplification can be usually categorized after doing literature review and reading previous publications on text simplification."\newline
\# Category: direct verbatim edit\newline
\# Explanation: This category of instruction requests to directly add information that are provided in the instruction verbatim to the answer.\newline
\newline
Example 10\newline
\# Original question: How have contrastive learning techniques been applied to learn dense sentence representations?\newline
\# Instruction: paragraph 3 does not describe another approach, \newline
\# Category: others\newline
\# Explanation: This category of instruction doesn't belong to any of the category above.\newline
\newline
Now, assign a category for the example below. ONLY ASSIGN ONE OF THE CATEGORY ABOVE. DONOT INVENT YOUR OWN CATEGORY. DON'T PROVIDE AN EXPLANATION. \\
\bottomrule
\end{tabular}
}
\caption{Prompt for automatic instruction analysis.}
\label{tab:instruction_analysis_prompt}
\end{table*}

\begin{table*}[]
\scriptsize
\resizebox{\textwidth}{!}{%
\begin{tabular}{@{}p{15cm}@{}}
\toprule
You are given an instruction for a model to improve an answer to a question and a comment on how the updated answer followed the instruction. Your job is to assign the comment into one of the category below. \newline
\newline
Example 1\newline
\# Instruction: Now add this information: ""[2] introduce a controllable summarization model that provides a mechanism for users to specify high level attributes such as length, style, or entities of interest. This enables personalized generation.""\newline
\# Comment: It added the information, but in the wrong place, so it looks as an extension of another method.
\# Category: added information but made answer less coherent/correct\newline
\# Explanation: This category of comment will mention that the instruction is followed, but the updated answer is less coherent (for example, become too long; has a weird structure, or is not correct). \newline
In the example, the comment mentioned that the updated answer has the information added but the updated answer is confusing.\newline
\newline
Example 2\newline
\# Instruction: Remove the tenth sentence.\newline
\# Comment: The sentence was not removed as requested.\newline
\# Category: instruction is ignored\newline
\# Explanation: This category of comment will mention that the instruction is not followed at all. In this example, the comment mentioned that requested change was not performed.\newline
\newline
Example 3\newline
\# Instruction: As a next sentence after this one "These parameters include the weights and biases of the model’s layers, such as the embedding layer, encoder, decoder, and attention mechanisms [0]." add a short sentence, which explains why parameters sharing is necessary in multilingual models used for machine translation\newline
\# Comment: It added an explanation, but deleted a whole paragraph\newline
\# Category: introduced unrequested changes\newline
\# Explanation: This category of comment will mention that the instruction is followed but also there are unrequested changes performed. In this example, the comment mentioned that the requested change (explanation) was incorporated but it also introduced unrequested change (a whole paragraph was deleted).\newline
\newline
Example 4\newline
\# Instruction: Great answer! Without changing anything else, in the last paragraph, mention some tasks that benefit from using explanations, and other tasks where it impedes performance.\newline
\# Comment:  The answer is overall still good but the new material is too long, and the model disregarded the part of the prompt where it was asked to put this material in the last paragraph. I will try to re-prompt.\newline
\# Category: failed to follow hard constraint\newline
\# Explanation: This category of comment will mention one ore more hard constraint in the instruction is not followed by the model. \newline
The hard constraint can be a specific location (at the beginning of the answer), length of the added/final answer (a very short definition), or an action to avoid ("do not...").\newline
In this example, the comment mentioned that the updated answer didn't follow the constraint in the instruction (which mention the added information should be in the last paragraph).\newline
\newline
Example 5\newline
\# Instruction: I notice a lot of redundant sentences, such as repeated mention of something like "compress the model while maintaining performance." Can you make your answer more concise?\newline
\# Comment:  model mostly just added paragraphs together without cutting actual content\newline
\# Category: others\newline
\# Explanation: This category of comment will mention that the model tried to follow the instruction (thus not completely ignore the instruction) but the updated answer is not satisfactory.\newline
The updated answer doesn't has issue mentioned above (unrequested change or failed to follow hard constraint), but just in general has subpar quality.\newline
\newline
Now, assign a category for the example below. ONLY ASSIGN ONE OF THE CATEGORY ABOVE. DONOT INVENT YOUR OWN CATEGORY. DON'T PROVIDE AN EXPLANATION. \\
\bottomrule
\end{tabular}
}
\caption{Prompt for automatic comment analysis.}
\label{tab:comment_analysis_prompt}
\end{table*}

\end{document}